\documentclass[10pt,twocolumn,letterpaper]{article}

\usepackage{iccv}
\usepackage{subcaption}
\usepackage{times}
\usepackage{epsfig}
\usepackage{graphicx}
\usepackage{amsmath}
\usepackage{amssymb}

\usepackage{color}
\usepackage{siunitx}
\usepackage{multirow}
\usepackage{stfloats}
\usepackage{amsthm}
\usepackage{comment}
\usepackage{listings}
\usepackage{algorithm}
\usepackage[accsupp]{axessibility} 

\usepackage{wrapfig}
\usepackage{epsfig}
\usepackage[utf8]{inputenc} 
\usepackage[T1]{fontenc}    
\usepackage{siunitx}
\usepackage{cite}
\usepackage{ctable}
\usepackage{hhline}
\usepackage{booktabs}

\usepackage{csquotes}

\usepackage[pagebackref=true,breaklinks=true,letterpaper=true,colorlinks,bookmarks=false]{hyperref}

\usepackage{color, colortbl}
\definecolor{Gray}{gray}{0.9}
\definecolor{Red}{RGB}{230, 57, 70}

\newlength\savewidth\newcommand\shline{\noalign{\global\savewidth\arrayrulewidth
  \global\arrayrulewidth 1pt}\hline\noalign{\global\arrayrulewidth\savewidth}}
  
\def\iccvPaperID{2268} 

\newcommand{\modelname}{\emph{Click-Pose}\xspace}
\iccvfinalcopy 
\ificcvfinal\fi

\usepackage{url}

\usepackage{capt-of,etoolbox}

\makeatletter
\apptocmd\@maketitle{{\myfigure{}\par}}{}{}
\makeatother

\begin{document}

\title{Neural Interactive Keypoint Detection}

\author{Jie Yang$^{1,2}$\thanks{Work done during an internship at IDEA.}~~, Ailing Zeng$^{1}$\thanks{Corresponding author.}~~, Feng Li$^{1}$, Shilong Liu$^{1}$, Ruimao Zhang$^{2}$\footnote[2]{}~~, Lei Zhang$^{1}$ \\
$^{1}$International Digital Economy Academy\\
$^{2}$School of Data Science, Shenzhen Research Institute of Big Data, \\The Chinese University of Hong Kong, Shenzhen \\
\texttt{\small{\{jieyang5@link,zhangruimao@\}cuhk.edu.cn}}\\
\texttt{\small{\{zengailing,lifeng,liushilong,leizhang\}@idea.edu.cn}}\\
\url{https://github.com/IDEA-Research/Click-Pose}
}

\ificcvfinal\thispagestyle{empty}\fi

\newcommand\myfigure{%
\vspace{-0.9cm}
\centering
\includegraphics[width=1.0\linewidth,trim={4pt 4pt 4pt 4pt}]{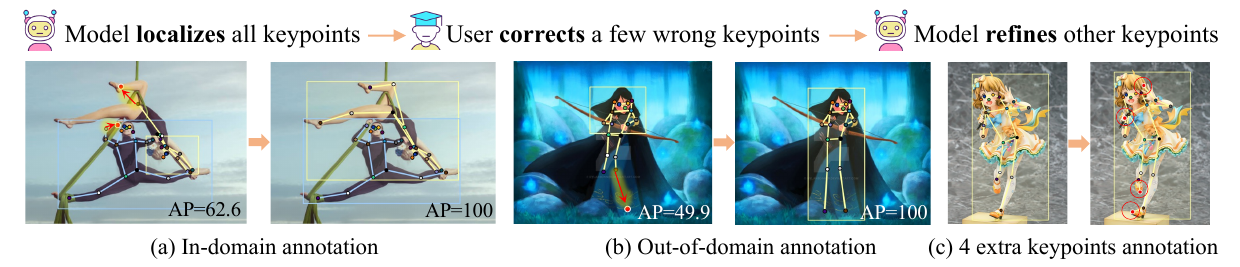}
\vspace{-0.5cm}
\captionof{figure}
{
We demonstrate the effects of \modelname in three keypoint annotation scenarios. For scenarios (a) and (b), the left figures show the model-only's initial keypoint localization, followed by the corrected keypoints (red points) obtained through user clicks.
The right figures display the final results obtained by \modelname after automatically refining other keypoints and corresponding human boxes. For scenario (c), the left figure illustrates the original task of detecting $17$ keypoints, while the right figure shows the adaptability of \modelname in detecting additional $4$ keypoints.
}
\label{fig:teaser}
\vspace{1.0em}
}

\maketitle
\begin{abstract}
This work proposes an end-to-end neural interactive keypoint detection framework named Click-Pose, which can significantly reduce more than 10 times labeling costs of 2D keypoint annotation compared with manual-only annotation.
Click-Pose explores how user feedback can cooperate with a neural keypoint detector to correct the predicted keypoints in an interactive way for a faster and more effective annotation process. 
Specifically, we design the pose error modeling strategy that inputs the ground truth pose combined with four typical pose errors into the decoder and trains the model to reconstruct the correct poses, which enhances the self-correction ability of the model.
Then, we attach an interactive human-feedback loop that allows receiving users' clicks to correct one or several predicted keypoints and iteratively utilizes the decoder to update all other keypoints with a minimum number of clicks (NoC) for efficient annotation.
We validate Click-Pose in in-domain, out-of-domain scenes, and a new task of keypoint adaptation. 
For annotation, Click-Pose only needs 1.97 and 6.45 NoC@95 (at precision 95\%) on COCO and Human-Art, reducing 31.4\% and 36.3\% efforts than the SOTA model (ViTPose) with manual correction, respectively. 
Besides, without user clicks, Click-Pose surpasses the previous end-to-end model by 1.4 AP on COCO and 3.0 AP on Human-Art. 

\end{abstract}
\vspace{-0.4cm}
\section{Introduction}

\begin{figure}[h]
\begin{center}
    \includegraphics[width=0.4\textwidth]{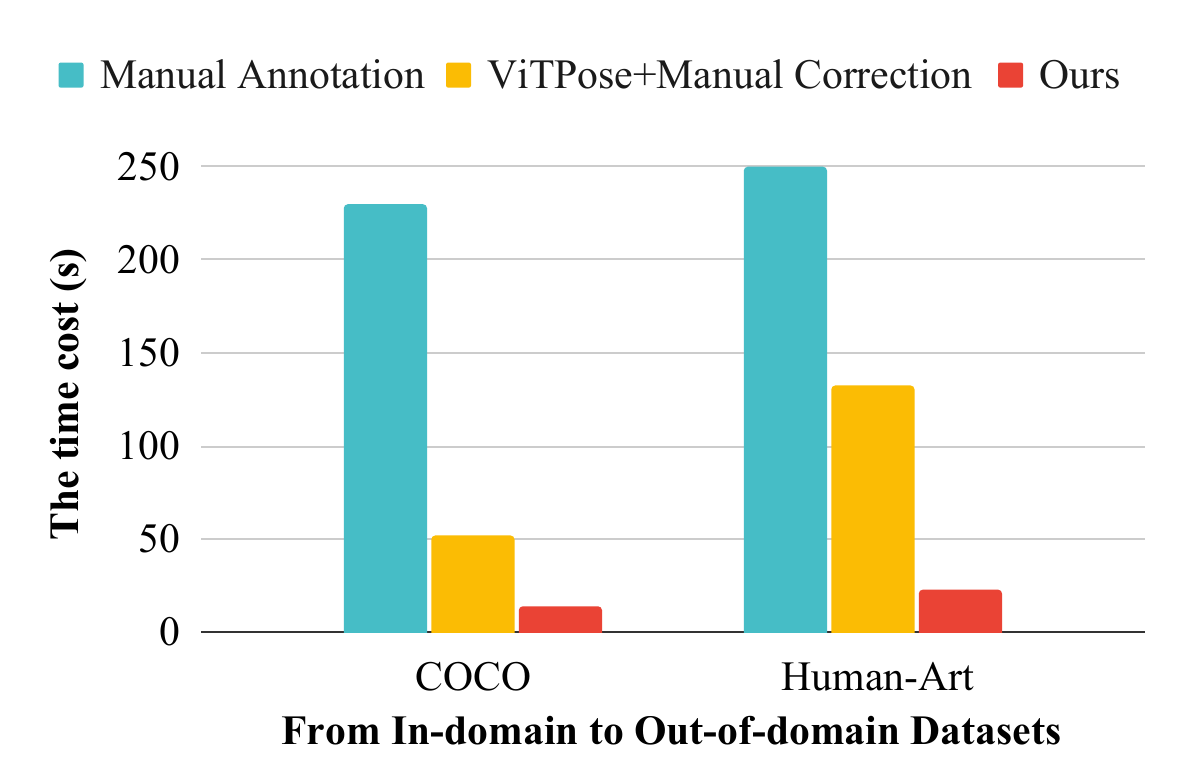}
\end{center}
\vspace{-0.5cm}
\caption{
Comparison of the average time cost of keypoint annotation per person using three strategies on two datasets. Our proposed \modelname is more than \textbf{10} times faster than manual annotation. Importantly, it significantly alleviates model bias in out-of-domain annotation (e.g., on Human-Art), reducing the time required by \textbf{83}\% compared to state-of-the-art model annotation with manual correction.}
\label{figure:fig1}
\vspace{-0.5cm}
\end{figure}

Multi-person keypoint detection aims to localize 2D coordinates of keypoints for each person in images, as in Fig.~\ref{fig:teaser}. It has garnered significant attention in research and industry, particularly in sports, entertainment, and surveillance applications. 
The development of deep models for various applications heavily depends on a large volume of training data with labels (e.g., COCO~\cite{Lin2014Microsoftcoco,Jin2020WholeBodyHP}). As the amount of data increases, the manual annotations of dense human keypoints are quite time-consuming, labor-intensive, and cost-prohibitive.
As demonstrated in Fig.~\ref{figure:fig1}, 
annotating a single person with $17$ keypoints would take about $230$ seconds. For a dataset of $50K$ images with an average of four people per image, this process would require $532$ hours.
Additionally, there may exist omissions, localization deviation, and mislabeling
in the manual annotation process. 

To reduce the manual effort, an intuitive annotation process can use a state-of-the-art (SOTA) model~\cite{xu2022vitpose} to obtain a preliminary model-annotation result and then mannually correct all wrong keypoints.
However, this strategy heavily relies on the performance of the model to reduce manual effort, which leads to the following problems:
1) \textbf{model bias}. As shown in Fig.~\ref{figure:fig1}, for in-domain data, the SOTA model (e.g., ViTPose-H) can accelerate the annotation process by about four times due to its high prediction accuracy. However, its performance may be suboptimal when labeling an out-of-distribution (OOD) dataset (e.g., Human-Art~\cite{humanart}) or when dealing with new keypoints~\cite{Xu2022PoseFE} that have not been defined. In such cases, more manual efforts will be required.
2) \textbf{performance bottleneck}. 
The performance of existing SOTA models is generally hard to be further improved,
which makes it challenging to further reduce manual effort. 
%
%
Noticing that there exist inherent problems in both the manual-only annotation and the model with manual correction strategies, the following questions naturally arise: \emph{how can we integrate manual correction with model predictions in an interactive manner to enable faster, more accurate, and more versatile keypoint annotation with minimal user correction?}

To address this issue, we define a novel task called interactive keypoint detection. It aims to effectively maximize benefits of the model to minimize manual effort, and mitigate unfriendly consequences of model failures in out-of-distribution and new-task annotations that increase the need for manual intervention. 
%
%
Accordingly, we present the first neural interactive keypoint detection framework, \modelname, as a baseline for further research. It allows a user to direct correct the positions of one or multiple keypoints and incorporate this feedback to refine other keypoints in Fig.~\ref{fig:teaser}.
%

Specifically, we build \modelname upon the end-to-end SOTA model ED-Pose~\cite{yang2023edpose}. This model incorporates a keypoint-to-keypoint refinement scheme through a regression head and updates keypoints layer-by-layer in the decoder, which allows receiving user-corrected positions at the decoder instead of the input image.
%
However, we empirically find that the decoder in ED-Pose is extremely susceptible to variations in input keypoint positions. Even a minor deviation can result in a significant deterioration in performance.
%
To tackle this limitation, we introduce two unique technical contributions to its decoder. The first is the pose error modeling that builds a reconstruction task 
to enhance the robustness of the decoder and learn to refine wrong keypoints by leveraging the correct keypoints as a reference. The second is the interactive human-feedback loop, which allows receiving users’ clicks to correct one or several predicted keypoints and iteratively utilizes the decoder to update all other keypoints with minimal manual corrections for efficient annotation.


\modelname incorporates the above two essential designs into the training process, which improves $+1.4$ AP on COCO \texttt{val} and $+3.0$ AP on HumanArt \texttt{val} compared with the baseline model ED-Pose, achieving state-of-the-art performance for end-to-end keypoint detection. More importantly, as shown in Fig.~\ref{fig:teaser}, \modelname shows its advantages in various annotation scenarios, \textit{i.e.,} in-domain, out-of-domain scenes, and a new task of keypoint adaptation. Specifically, \modelname only needs $1.97$ and $6.45$ NoC@95 (the average number of user clicks needed to annotate one person to achieve a precision of $95\%$) on COCO and Human-Art, reducing $31.4\%$ and $36.3\%$ efforts than the SOTA model with manual correction, respectively. Moreover, \modelname significantly reduces the average time cost of single-person annotation, achieving over $5\times$ speedup compared to the SOTA model ViTPose with manual correction and more than a $10\times$ speedup compared to manual-only annotation, especially in out-of-domain scenarios.

Our contributions are:
(1) We define a novel task called interactive
keypoint detection to pursue high-precision and low-cost annotation,
and present the first framework to address this task, namely \modelname. 
(2) We incorporate the pose error modeling and interactive human-feedback loop into the training of \modelname, leading to a state-of-the-art performance for end-to-end keypoint detection.
(3) We provide a new metric (NoC) and extensively validate the effectiveness and efficiency of \modelname in different annotation scenes.
%
We hope this work could inspire further research in related fields.
\section{Related work}

\subsection{Multi-Person Pose Estimation}
Existing pose estimation models can be generally divided into two-stage methods and one-stage methods. For two-stage methods, there are top-down (TD) and bottom-up (BU) strategies.
Top-down methods \cite{xu2022vitpose,mao2022poseur,xiao2018simple,sun2019deep,li2021pose} have achieved high performance by first detecting each person in the image with an object detector and then conducting the single-person pose estimation with the proposed model. However, these methods are limited by their inability to handle missing person detections and their high costs for crowd scenes. In contrast, bottom-up methods \cite{newell2017associative,cao2017realtime,geng2021bottom,cheng2020higherhrnet,luo2021rethinking} have demonstrated greater efficiency by first estimating keypoints and then grouping them into individual human poses. 
Recently, the advent of end-to-end object detectors DETR~\cite{carion2020end} has led to the development of one-stage pose estimators, like PETR~\cite{shi2022end} and ED-Pose~\cite{yang2023edpose}. ED-Pose is particularly notable in that it approaches this task as two explicit box detection processes, leading to superior performance and efficiency trade-offs.
Additionally, some refinement models \cite{moon2019posefix,kan2022self,zeng2022smoothnet,zeng2022deciwatch} also focus on pose correction. They take both the original image and an estimated pose as inputs to refine a more accurate pose. However, despite great efforts to achieve state-of-the-art models, such as ViTPose~\cite{xu2022vitpose} with ViT-Huge backbone~\cite{dosovitskiy2020image}, and other models~\cite{moon2019posefix,kan2022self} that specialize in pose refinement, they still require manual correction to satisfy the precision requirement of annotation, where even more manual effort is required to compensate for the performance drop in out-of-distribution annotation scenarios.
In contrast, we attempt to address interactive keypoint annotation with minimal manual efforts using a fully end-to-end framework.




\subsection{Human-in-the-Loop}
Annotation is a typical application scenario for Human-in-the-Loop (HITL) techniques, which aims to improve prediction models' accuracy while minimizing costs by leveraging human knowledge and experience. Existing works mainly focus on two directions: (i) data processing via human feedback. For instance, one approach, known as \textit{active learning}~\cite{aghdam2019active,kim2021task}, seeks to minimize manual annotation effort on a large dataset while maximizing the model's performance~\cite{xie2022towards,shukla2022vl4pose,feng2023rethinking}; 
In specific, prior HITL methods for pose estimation~\cite{liu2017active, gong2022meta, feng2023rethinking} have involved actively selecting and labeling informative images to facilitate effective learning.  
(ii) interventional model training and inference via human feedback. 
For instance, the interactive image segmentation task is to extract an accurate target mask with minimal user interaction~\cite{Xu2016DeepIO,sofiiuk2020f,chen2022focalclick,liu2022simpleclick}. This is a popular research area over the past years. Existing deep learning-based approaches usually input both the image and user annotations in the model training and testing stages or conduct various inference-time optimization schemes~\cite{jang2019interactive,kontogianni2020continuous}, which suffer from high computation costs and slow speeds for each annotation. A prior study~\cite{kim2022morphology} introduces an interactive image segmentation pipeline designed for heatmap-based interactive keypoint annotation in X-ray images. Additionally, researcher\cite{cormier2021interactive} has delved into the realm of inter- and extrapolated annotations across frames. Remarkably, 
\emph{no work has yet explored how to enable effective interaction between deep models and human feedback to improve end-to-end multi-person keypoint annotation accuracy with fewer costs and manual efforts}.
In this work, inspired by the concept of HITL, we first investigate how to combine human feedback with a deep model in an interactive manner for human body keypoint annotation.

\section{Methodology}
\begin{figure*}[h]
    \centering
    \includegraphics[width=1\linewidth]{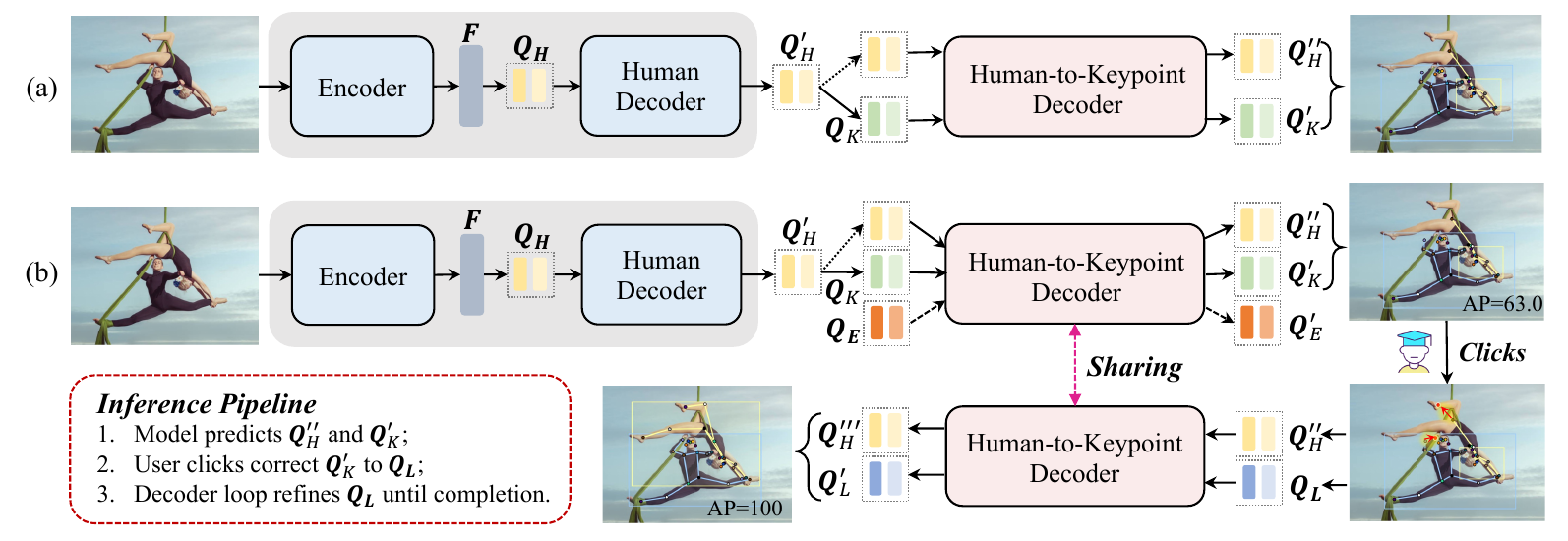}
    \caption{Comparison of (a) ED-Pose~\cite{yang2023edpose}  with (b) the proposed \modelname. 
    \modelname follows the same architecture as ED-Pose while introducing two key techniques to the Human-to-Keypoint decoder.
    Firstly, we introduce a \textit{\textbf{training-only}} strategy, namely \textbf{Pose Error Modeling}. It builds a reconstruction task to self-correct error-keypoint queries $\textbf{Q}_E$ to $\textbf{Q}_E^{\prime}$, which enhances the robustness of the model and learns to refine wrong keypoints by leveraging correct keypoints as a reference. Secondly, we attach an \textbf{Interactive Human-Feedback Loop} to allow the user to correct one or several keypoints positions in $\textbf{Q}_K^{\prime}$ and generate the modified keypoint queries $\textbf{Q}_L$. Then the Human-to-Keypoint decoder could take the predicted boxes $\textbf{Q}_H^{\prime \prime}$ and $\textbf{Q}_L$ as input again and further refine human boxes and all keypoints to $\textbf{Q}_H^{\prime \prime \prime}$ and $\textbf{Q}_L^{\prime}$ based on user corrections.
    }
    \label{fig:overview}
    \vspace{-0.3cm}
\end{figure*}

\subsection{Motivation}

\textbf{Interactive Keypoint Detection.} Interactive keypoint detection aims to obtain accurate keypoint annotations with minimal user interactions. For example, if a network predicts an incorrect pose, such as a flipped pose, the user may only need to correct one keypoint by clicking on it. Subsequently, the network can use this human feedback to further refine the remaining keypoint positions and determine the correct pose.
To address this task, the network should incorporate a pose-to-pose refinement scheme that can receive modified keypoint positions from the user and output further refined positions. 

\textbf{Preliminary Study of ED-Pose~\cite{yang2023edpose}.} ED-Pose addresses the task of keypoint detection by explicitly reformulating $4$D keypoint boxes as queries and progressively refining them layer by layer in the decoder through a regression head. It achieves SOTA performance compared with existing end-to-end models and improves the inference speed. When considering only the $2$D coordinate, ED-Pose can be seen as providing a keypoint-to-keypoint refinement scheme, thus conceptually satisfying the aforementioned architectural requirement for interactive keypoint detection.
As shown in Fig.~\ref{fig:overview}-(a), it consists of an Encoder, a Human decoder, and a Human-to-Keypoint decoder. Specifically, it extracts image features using a backbone and passes them through the Encoder with positional embedding to obtain refined image features $\textbf{F}$. In the Human decoder, ED-Pose leverages human queries $\textbf{Q}_H$ to search for human objects, where $\textbf{Q}_H$ contains position queries $\textbf{Q}_H^p$ (\textit{i.e.,} human box positions) and content queries $\textbf{Q}_H^c$ (\textit{i.e.,} human content embedding). Then, it utilizes the updated human queries $\textbf{Q}_H^{\prime}$ to initialize keypoint queries $\textbf{Q}_K$, where $\textbf{Q}_K$ also includes position queries $\textbf{Q}_K^p$ (\textit{i.e.,} keypoint positions) and content queries $\textbf{Q}_K^c$ (\textit{i.e.,} keypoint content embedding). Finally, it attaches the Human-to-keypoint decoder to refine the human box and keypoints of each person to $\textbf{Q}_H^{\prime \prime}$ and $\textbf{Q}_K^{\prime}$.

\textbf{Non-interactive Issue in ED-Pose.}
As mentioned above, the Human-to-Keypoint decoder of ED-Pose can be viewed as a keypoint-to-keypoint refinement process. It is natural to consider whether human feedback (e.g., a corrected keypoint) can be directly incorporated 
in the decoder without any further modification.
However, extensive preliminary experiments have shown that the decoder is highly sensitive to the input keypoint position query $\textbf{Q}_K^p$ and is unable to effectively utilize human feedback. For example, 
during the inference, we randomly add a small disturbance $(\Delta x,\Delta y)$ to each keypoint coordinate $(x,y)$ in $\textbf{Q}_K^p$. We ensure that $\lvert \Delta x\rvert<\omega_x$ and $\lvert \Delta y\rvert<\omega_y $, where $\omega_x, \omega_y \in (0,0.1)$. This operation results in a sharp drop in accuracy from $71.6$AP to $11.8$AP. 
There are two main reasons for this sensitivity: Firstly, the Human-to-Keypoint decoder effectively learns the contextual information of each keypoint, which leads to a strong coupling between the position query $\textbf{Q}_K^p$ and the content query $\textbf{Q}_K^c$. Once $\textbf{Q}_K^p$ changes, this misalignment can cause the final results to drop off. Secondly, the Human-to-Keypoint decoder also creates a contextual coupling relationship among different keypoints, meaning that adjusting the input position of one keypoint may severely harm the update of the others.

\subsection{The Overview of Click-Pose}
\textbf{Introduction to \modelname.} As illustrated in Fig.~\ref{fig:overview}-(b), \modelname adopts the same modules as ED-Pose to obtain the person box $\textbf{Q}_H^{\prime \prime}$ and keypoints $\textbf{Q}_K^{\prime}$ from an input image. Furthermore, \modelname introduces two key techniques to the Human-to-Keypoint Decoder, which makes the model interactive and robust.
Firstly, we additionally introduce error-keypoint queries $\textbf{Q}_E$ in the training stage, which includes four typical pose errors defined by~\cite{ruggero2017benchmarking}. We feed them into the Human-to-Keypoint decoder to reconstruct the accurate pose $\textbf{Q}_E^{\prime}$. This operation enhances the self-correction ability of the Human-to-Keypoint decoder. We call it \textbf{\textit{Pose Error Modeling}} (see Sec.~\ref{sec:error}).
Secondly, we introduce the user interaction in an attached Human-to-Keypoint decoder via proposed \textbf{\textit{Interactive Human-Feedback Loop}} (see Sec.~\ref{sec:loop}). In this process, the user can correct one or several keypoints positions in $\textbf{Q}_K^{\prime}$ and generate the modified keypoint queries $\textbf{Q}_L$. Then, the Human-to-Keypoint decoder could take $\textbf{Q}_H^{\prime \prime}$ and $\textbf{Q}_L$ as input iteratively and further refine human boxes to $\textbf{Q}_H^{\prime \prime \prime}$ and all keypoints to $\textbf{Q}_L^{\prime}$ based on the modified keypoints. This process can further improve the self-correction ability of the model during the training and successfully allow the user clicks to be integrated into the inference phase.

\textbf{Training Optimization Processes of \modelname.}
\modelname is an end-to-end trainable framework that extends the ED-Pose training process (the loss as $\mathcal{L}_{g}$). Firstly, \textit{Pose Error Modeling} uses ground-truth keypoints to generate erroneous poses and creates a pose reconstruction task, introducing the loss as $\mathcal{L}_{r}$.
Secondly, \textit{Interactive Human-Feedback Loop} uses the ground-truth keypoints to simulate user clicks for correcting a few wrong keypoints in model predictions and  loop decoder to refine other wrong keypoints and the corresponding human boxes, introducing the loss as $\mathcal{L}_{l}$. 
Finally, the overall training pipeline of \modelname can be written as follows,
\begin{equation}
    \mathcal{L} = \mathcal{L}_g+ \mathcal{L}_r + \mathcal{L}_l,
\end{equation}
where we employ a set-based Hungarian matching to ensure a unique prediction for each ground-truth pose~\cite{yang2023edpose,carion2020end,shi2022end}. Following ED-Pose, $\mathcal{L}_g$ and $\mathcal{L}_l$ include human classification loss, human box regression loss, and human pose regression loss. $\mathcal{L}_r$ is the $L1$ loss for pose reconstruction.

\textbf{Inference Pipeline of \modelname.}
Given an image, \modelname first performs an end-to-end inference to obtain all human boxes $\textbf{Q}_H^{\prime \prime}$ and keypoint locations $\textbf{Q}_K^{\prime}$ without any troublesome post-processing. Notably, $\textbf{Q}_E$ is not required in inference. In annotation scenes, the user can directly correct the wrong keypoints in model predictions $\textbf{Q}_K^{\prime}$ to $\textbf{Q}_L$ and loop decoder to refine human boxes $\textbf{Q}_H^{\prime \prime \prime}$ and all keypoints $\textbf{Q}_L$ until the annotation is completed.

\subsection{Pose Error Modeling}
\label{sec:error}
 Pose Error Modeling aims to enhance the robustness of the Human-to-Keypoint decoder via a reconstruction scheme~\cite{li2022dn}.
To achieve this, we introduce error-keypoint queries $\textbf{Q}_E$
 by adding four typical error types into ground-truth keypoints, and then we feed $\textbf{Q}_E$ into the Human-to-Keypoint decoder to reconstruct the accurate poses $\textbf{Q}_E^{\prime}$.
 
In specific, $\textbf{Q}_E$ consists of position queries $\textbf{Q}_E^p$ and content queries $\textbf{Q}_E^c$, where the former can be initialized by the 2D coordinates of keypoints and the latter can be initialized by the keypoint label embedding via a learnable codebook $\textbf{B} \in \mathbb{R}^{ K \times C}$. $K$ is the number of defined keypoints, and $C$ is the channel dimension. Then, we simulate four typical error types of the keypoint, \textit{i.e.,} \textit{jitter}, \textit{miss}, \textit{swap} and \textit{inversion} defined by~\cite{ruggero2017benchmarking,moon2019posefix}, and add them into the ground-truth keypoints for the initialization of $\textbf{Q}_E^p$ and $\textbf{Q}_E^c$.

For localization issues, \textit{i.e.,} \textit{jitter}, \textit{miss}, and \textit{swap},
we perturb the ground-truth keypoints with different magnitudes of position disturbance to initial $\textbf{Q}_E^p$. Specifically, we add a random disturbance $(\Delta x,\Delta y)$ to the $(x,y)$ of the keypoint and make sure that $\lvert \Delta x\rvert<\frac{\lambda_x w}{2}$ and $\lvert \Delta y\rvert<\frac{\lambda_y h}{2}$, where $\lambda_x, \lambda_y \in (0,1)$. Such disturbance constrains the keypoints with pose errors to remain within the bounding box. In addition, we use $\textbf{B}$ directly to embed ground-truth keypoint labels to initialize the $\textbf{Q}_E^c$.
Moreover, \textit{inversion} is a complex error that involves mislabeling and mislocating body parts within the same person (e.g., confusing the left and right elbow). 
As perturbing positions of ground-truth keypoints to initialize $\textbf{Q}_E^p$, we have a hyper-parameter $\alpha$ (e.g., $0.4$) to randomly flip the labels of the left and right body parts for the initialization of $\textbf{Q}_E^c$. Such keypoint flipping introduces a misalignment between $\textbf{Q}_E^p$ and $\textbf{Q}_E^c$, which compels the model to recognize the interdependence between the position and label of the body parts.
At last, we preserve a subset of ground-truth keypoints in $\textbf{Q}_E$. This enables the model to learn how to leverage the correct ground-truth keypoints as a reference to refine wrong keypoints.

\subsection{Interactive Human-Feedback Loop}
\label{sec:loop}
Interactive Human-Feedback Loop aims to interact with user clicks, minimize manual corrections and enable efficient annotation. It allows the model to receive user clicks for correcting a few predicted keypoints and iteratively utilize the proposed decoder to update all other keypoints and human boxes.

\textbf{Initialization of Modified Queries $\textbf{Q}_L$.}
Given the predicted keypoints $\textbf{Q}_K^{\prime}$, which contains position queries $\textbf{Q}_K^{q\prime}$ and content queries $\textbf{Q}_K^{c\prime}$, the user can click the one or several keypoints in $\textbf{Q}_K^{q\prime}$ to obtain the modified position queries $\textbf{Q}_L^p$ of $\textbf{Q}_L$, where $\textbf{Q}_L^p$ only have a few keypoints corrected by the user (e.g., $1$ click). Since the modified position queries $\textbf{Q}_L^p$ and the originally predicted content queries $\textbf{Q}_K^{c\prime}$ are misaligned, we initialize the modified content queries $\textbf{Q}_L^c$ through label embedding using the codebook $\textbf{B}$
, which is shared with the pose error modeling process.

\textbf{Training and Inference Strategies.}
For training, we employ Hungarian matching to obtain the predicted poses that are matched with ground-truth poses in an image. Then, we can directly modify the corresponding predicted keypoints $\textbf{Q}_K^{q\prime}$ using the ground-truth to simulate the user click operation and obtain the modified queries $\textbf{Q}_L$. We loop decoder to refine human boxes $\textbf{Q}_H^{\prime \prime}$ and all keypoints $\textbf{Q}_L$ to $\textbf{Q}_H^{\prime \prime \prime}$ and $\textbf{Q}_L{\prime}$, which are supervised by ground-truth boxes and keypoints.
For inference, to obtain quantitative results, we evaluate the effectiveness and efficiency of \modelname for annotation on existing datasets in a manner similar to the training procedure. In real annotation scenarios, \modelname enables users to provide direct feedback to complete annotation with minimal effort.

\section{Experiments}

\subsection{Experimental Setup}
\textbf{Datasets.}
We evaluate our methods on four benchmarks: COCO~\cite{Lin2014Microsoftcoco}, Human-Art~\cite{humanart}, OCHuman~\cite{zhang2019pose2seg} and CrowdPose~\cite{li2019crowdpose}. COCO consists of about 250K person instances with $17$ keypoints, and provides diverse human poses in natural scenarios. On the other hand, Human-Art comprises 123K person instances with $21$ keypoints, of which $17$ are the same as COCO. It provides rich human poses in out-of-distribution artistic scenes. OCHuman has $8110$ human pose instances that have occlusions with the maxIOU$\geq$$0.5$, where $32\%$
instances are more challenging with the maxIOU$\geq$$0.75$. CrowdPose provides $80000$ human poses with $14$ labeled keypoints in the crowded scenes.

\textbf{Evaluation Metrics.}
Inspired by interactive segmentation~\cite{sofiiuk2020f,liu2022simpleclick}, we introduce a new metric called the Number of Clicks (NoC), which measures the average number of clicks needed to annotate one person to achieve a specific target average precision (AP). We set the target AP to 85\%, 90\%, and 95\%, denoting the corresponding measures as NoC$@85$, NoC$@90$, and NoC$@95$, respectively. The average NoC is calculated over images that contain person instances in the COCO \texttt{val} set or Human-Art \texttt{val} set for evaluation. Moreover, we report the overall AP when restricting the number of clicks per person, such as C$1$ and C$3$, which aims to evaluate the performance that different methods can achieve with the same human effort.

\begin{table*}[h]
\begin{center}
    \begin{minipage}{0.41\linewidth}
        \begin{center}
            \scriptsize
            \resizebox{\linewidth}{!}{
                \makeatletter\def\@captype{table}\makeatother
                \begin{tabular}{c|c|c|c}
                    \shline
                    Time cost (s) & Manual-only & ViTPose+C & Ours \\ \hline
                    COCO & 230$\pm$56 & 52$\pm$10 & \textbf{14$\pm$5} \\ 
                    Human-Art & 250$\pm$55 & 132$\pm$41 & \textbf{23$\pm$8} \\ 
                    \shline
                \end{tabular}}
        \end{center}
        \vspace{-0.5cm}
        \caption{Comparisons of \textbf{the average and standard deviation time cost} required for single-person annotation by three annotation strategies: manual-only, SOTA model with manual correction, and our \modelname.
        }
        \vspace{-0.5cm}
        \label{tab:user}
    \end{minipage}
        \hspace{0.3cm}
    \begin{minipage}{0.55\linewidth}
        \begin{center}
            \scriptsize
            \resizebox{\linewidth}{!}{
                \makeatletter\def\@captype{table}\makeatother
                \begin{tabular}{l|c|ccc}
                    Method  & Backbone &  NoC$@85\downarrow$ & NoC$@90\downarrow$ & NoC$@95\downarrow$  \\
                    \shline
                    \multicolumn{5}{l}{\textit{\cellcolor{Gray}COCO \texttt{val}}} \\
                    ViTPose & ViT-Huge & 1.46  & 2.15 & 2.87\\        
                    \modelname & ResNet-50 & \textbf{0.95} & \textbf{1.48} & \textbf{1.97}\\
                    \shline
                    \multicolumn{5}{l}{\textit{\cellcolor{Gray}Human-Art \texttt{val} }} \\
                    ViTPose & ViT-Huge  &  9.12 &9.79 &10.13\\ 
                    \modelname & ResNet-50  & \textbf{4.82} & \textbf{5.81} & \textbf{6.45}\\ 
                    \bottomrule
                \end{tabular}}
        \end{center}
        \vspace{-0.5cm}
        \caption{Comparisons of \textbf{Number of Clicks (NoC)} metrics for interactive keypoint detection.}
        \label{tab:NoC}
    \end{minipage}
    \end{center}
        \vspace{-0.8cm}
\end{table*}

\textbf{Implementation Details.} 
Following~\cite{yang2023edpose,carion2020end}, the training images are augmented by random cropping, flipping, and resizing with the shorter sides in $[480, 800]$ and the longer sides less or equal to 1333. The number of queries $\textbf{Q}_K$ is set to $50$.
We use the AdamW optimizer with a weight decay of $1\times10^{-4}$. Our model is trained on Nvidia A100 GPUs with a batch size of $16$ for $40$ epochs on COCO. The initial learning rate is $1\times10^{-4}$ and is reduced by a factor of $0.1$ at the 38th epoch on COCO. The channel dimension C is set to $256$. The testing images are resized to have shorter sides of 800 and longer sides less than or equal to 1333. All compared DETR-based models use the ResNet-50 backbone.

\begin{table*}[h]
    \begin{center}
    \scriptsize
        \begin{tabular}{l|c|ccccc|c}
            Method & Backbone & AP$\uparrow$ & AP$_{50}$$\uparrow$ & AP$_{75}$$\uparrow$ & AP$_{M}$$\uparrow$ & AP$_{L}$$\uparrow$ & Time [ms]$\downarrow$\\
            \shline
            \multicolumn{8}{l}{\textit{\cellcolor{Gray}Model-Only}} \\
            ViTPose$^{\dag}$~\cite{xu2022vitpose} (TP) &  ViT-Huge & 79.1 & 91.7& 85.7 & 71.9 & 82.0 & 45+286 \\
            HRNet$^{\dag}$~\cite{sun2019hrnet} (TP) & HRNet-w32 & 74.4 & 90.5 & 81.9 & 70.8 & 81.0 & 45+112 \\
            HrHRNet$^{\dag}$~\cite{cheng2020higherhrnet} (BU) & HRNet-w32 & 67.1 & 86.2 & 73.0 & 61.5 & 76.1 & 322 \\
            PETR~\cite{shi2022end} (OS) & ResNet-50  & 68.8 & 87.5 & 76.3 & 62.7 & 77.7 &105\\
            ED-Pose~\cite{yang2023edpose} (OS) & ResNet-50 &71.6 & 89.6 & 78.1 & 65.9 & 79.8 &51 \\
            \modelname-C0 (OS) & ResNet-50 & 73.0\color{Red}$\uparrow_{1.4}$ & 90.4 &80.0 & 68.1 & 80.5 & \textbf{48}\color{Red}$\downarrow_{3}$ \\
            \shline
            \multicolumn{8}{l}{\textit{\cellcolor{Gray}Model+Manual Correction}} \\
        ViTPose-C1 &  ViT-Huge & 82.3 & 90.8  & 86.6 & 78.8 & 87.9 & - \\
        ViTPose-C2 &  ViT-Huge  & 85.3 & 91.9  & 89.5 & 83.6 & 89.4 & -\\
        ViTPose-C3 &  ViT-Huge  & 86.7 & 93.8  & 90.3 & 86.7 & 89.4 & -\\
        ViTPose-C4 &  ViT-Huge  &  88.3 & 95.2 & 92.4 & 90.9 & 89.5 & -\\
            \shline
            \multicolumn{8}{l}{\textit{\cellcolor{Gray}Neural Interactive}} \\
            \modelname-C1 & ResNet-50 &  83.2 ({\color{Red}{+1.8}}) & 96.5 ({\color{Red}{+3.4}}) & 89.7 ({\color{Red}{+2.3}}) & 80.1 ({\color{Red}{+2.8}}) & 87.9 ({\color{Red}{+0.2}})  & - \\
            \modelname-C2 & ResNet-50 & 90.3 ({\color{Red}{+2.7}}) & 97.8 ({\color{Red}{+3.1}}) & 95.2 ({\color{Red}{+4.1}}) & 88.1 ({\color{Red}{+3.1}}) & 93.9 ({\color{Red}{+1.9}}) & -  \\
            \modelname-C3 & ResNet-50 & 94.1 ({\color{Red}{+3.4}}) & 98.9 ({\color{Red}{+3.4}}) & 96.6 ({\color{Red}{+3.8}}) & 92.6 ({\color{Red}{+3.5}}) & 96.5 ({\color{Red}{+2.8}}) & -\\
            \modelname-C4 & ResNet-50 & \textbf{96.4} ({\color{Red}{+3.9}}) & \textbf{99.0} ({\color{Red}{+3.3}}) & \textbf{97.9} ({\color{Red}{+4.3}}) & \textbf{95.3} ({\color{Red}{+3.8}}) & \textbf{97.8} ({\color{Red}{+3.3}}) &  - \\
            \shline
        \end{tabular}
    \end{center}
    \vspace{-0.4cm}
    \caption{\textbf{Comparison with representative SOTAs} on COCO \texttt{val} set. C1-C4 limits the number of clicks on a single person.
\modelname-C0 is a fully end-to-end framework without user clicks. The red arrow indicates its improvement over ED-Pose~\cite{yang2023edpose}. The number in parentheses is the interactive model improvement via the loop refinement (ignoring the manual improvement).
$\dag$ denotes the flipping test. The inference time of all model-only methods is tested on an A100, except for the detector of the top-down methods, which is referred from the MMdetection (\textit{i.e.,} 45ms).}
    \vspace{-0.2cm}
    \label{tab:coco_main}
\end{table*}

\begin{table}[h]
    \begin{center}
    \resizebox{\linewidth}{!}
    {%
        \begin{tabular}{l|ccccc|c}
            Method &  C0$\uparrow$ & C1$\uparrow$ & C2$\uparrow$ & C3$\uparrow$ & C4$\uparrow$ & NoC@95$\downarrow$ \\
                \shline
                        Poseur~\cite{mao2022poseur} (TP) &  74.2 & 80.9 & 84.8 & 86.4 & 88.6 & 3.15 \\ 
                        ED-Pose~\cite{yang2023edpose} (OS) & 71.6 & 80.1 & 84.4 & 86.9 & 88.5 & 5.40\\
            \modelname (OS) &    \textbf{73.0}  & \textbf{83.2} & \textbf{90.3} & \textbf{94.1} & \textbf{96.4} & \textbf{1.97}\\
            \shline
        \end{tabular}
    }
    \end{center}
    \vspace{-0.6cm}
    \caption{\textbf{Comparison with DETR-based models} on COCO \texttt{val} set.}
    \label{tab:COCO}
    \vspace{-0.3cm}
\end{table}

\begin{table}[h]
    \begin{center}
    \resizebox{\linewidth}{!}
    {%
        \begin{tabular}{l|c|ccc}
            Method & Backbone & AP & AP$_{M}$ & AP$_{L}$\\
            \shline
            \multicolumn{5}{l}{\textit{\cellcolor{Gray}Model-Only}} \\
            ViTPose (TP) &  ViT-Huge & 28.7 & 1.6 & 31.8  \\           
            HRNet (TP) & HRNet-w48  & 22.2 & 1.6 & 24.5 \\
            HrHRNet (BU) & HRNet-w48 & 34.6 & 5.6& 38.1 \\
            ED-Pose (OS) & ResNet-50  &  37.5 & 7.6 & 41.1\\
            \modelname-C0 (OS)   & ResNet-50   & 40.5{\color{Red}$\uparrow_{3.0}$} & 8.3 & 44.2 \\
            \shline
                    \multicolumn{5}{l}{\textit{\cellcolor{Gray}Model+Manual Correction}} \\
        ViTPose-C3 &  ViT-Huge &  32.1 & 5.1 & 34.8\\
        ViTPose-C5 &  ViT-Huge  &  36.1 & 12.3 & 38.3 \\
        ViTPose-C7 &  ViT-Huge  &  40.3 & 19.0 & 42.3\\
        ViTPose-C9 &  ViT-Huge  &  47.5 & 28.9 & 49.1 \\
                    \shline
            \multicolumn{5}{l}{\textit{\cellcolor{Gray}Neural Interactive}} \\
            \modelname-C3 & ResNet-50 & 61.6 ({\color{Red}{+13.4}})  & 30.8 ({\color{Red}{+16.7}}) & 65.1 ({\color{Red}{+13.3}})\\
            \modelname-C5 & ResNet-50 & 71.8 ({\color{Red}{+19.8}})  & 45.1 ({\color{Red}{+26.4}}) & 74.5 ({\color{Red}{+19.2}}) \\
            \modelname-C7 & ResNet-50 & 78.5 ({\color{Red}{+24.1}})  & 54.7 ({\color{Red}{+32.9}}) & 80.9 ({\color{Red}{+23.3}}) \\
            \modelname-C9 & ResNet-50 & \textbf{83.7} ({\color{Red}{+27.6}})  & \textbf{63.1} ({\color{Red}{+38.1}}) & \textbf{85.9} ({\color{Red}{+27.0}})  \\
            \shline
        \end{tabular}
    }
    \end{center}
    \vspace{-0.5cm}
    \caption{\textbf{Comparison with representative SOTAs} on Human-Art \texttt{val} set. All the models are trained on COCO and tested on Human-Art as out-of-distribution data.}
        \vspace{-0.2cm}
    \label{tab:Human-Art_main}
\end{table}

\begin{table}[h]
    \begin{center}
    \resizebox{\linewidth}{!}
    {%
        \begin{tabular}{l|ccccc|c}
            Method &  C0$\uparrow$ & C1$\uparrow$ & C2$\uparrow$ & C3$\uparrow$ & C4$\uparrow$ & NoC$@95$$\downarrow$ \\
                \shline
                        Poseur~\cite{mao2022poseur} (TP) &  21.2 & 23.1 & 24.9  & 26.4  & 28.0 & 12.19 \\ 
                        ED-Pose~\cite{yang2023edpose} (OS) & 37.5 & 40.1 & 42.0 &43.3   & 44.3 & 9.88 \\
            \modelname (OS) & \textbf{40.6} & \textbf{47.1}& \textbf{54.9} & \textbf{61.6}& \textbf{67.1} & \textbf{6.45}\\
            \shline
        \end{tabular}}
    \end{center}
    \vspace{-0.5cm}
    \caption{\textbf{Comparison with DETR-based models} on Human-Art \texttt{val} set.}
    \label{tab:Human-Art}
        \vspace{-0.5cm}
\end{table}

\begin{table}[h]
    \begin{center}
    \resizebox{\linewidth}{!}
    {%
        \begin{tabular}{l|cccc}
            Method & AP & AP$_{50}$ & AP$_{75}$ & NoC$@95$\\
            \shline
            \multicolumn{5}{l}{\textit{\cellcolor{Gray}Model-Only}} \\
            ED-Pose (OS)  & 31.4 & 39.5 & 35.1 & - \\
            \modelname-C0 (OS)   & 33.9{\color{Red}$\uparrow_{2.5}$} & 43.4 & 37.5 & - \\
            \shline
                    \multicolumn{5}{l}{\textit{\cellcolor{Gray}Model+Manual Correction}} \\
        ED-Pose-C1 &  33.0 & 39.6 & 35.5  & \multirow{2}{*}{{13.50}}\\
        ED-Pose-C2 &   33.7 & 39.6 & 35.6  \\
                    \shline
            \multicolumn{5}{l}{\textit{\cellcolor{Gray}Neural Interactive}} \\
            \modelname-C1 &  83.0 ({\color{Red}{+46.4}}) & 92.4 ({\color{Red}{+49.0}}) & 88.0 ({\color{Red}{+49.0}}) & \multirow{2}{*}{\textbf{1.93}}\\
            \modelname-C2 &  \textbf{90.9} ({\color{Red}{+52.4}}) & \textbf{96.3} ({\color{Red}{+52.5}}) & \textbf{93.3} ({\color{Red}{+53.4}}) \\
            \shline
        \end{tabular}
    }
    \end{center}
    \vspace{-0.5cm}
    \caption{\textbf{Comparison with baseline models} on the crowded scene, where all the models are trained on COCO and tested on OCHuman \texttt{test} set.}
    \label{tab:OCHuman}
\end{table}

\begin{table}[h]
    \vspace{-0.3cm}
    \begin{center}
    \resizebox{\linewidth}{!}
    {%
        \begin{tabular}{l|ccccc|c}
            Method &  C0$\uparrow$ & C1$\uparrow$ & C2$\uparrow$ & C3$\uparrow$ & C4$\uparrow$ & NoC$@95$$\downarrow$ \\
                \shline 
                        ED-Pose (OS) &   69.9  &77.6 &82.3 & 84.8 & 86.1 & 6.37\\
            \modelname (OS) &  \textbf{70.6} & \textbf{79.1} & \textbf{86.1} & \textbf{91.3} & \textbf{94.5} & \textbf{1.47} \\
            \shline
        \end{tabular}}
    \end{center}
    \vspace{-0.5cm}
    \caption{\textbf{Comparison with baseline models} on the crowded scene, where all the models are trained and tested on CrowdPose with defined $14$ keypoints.}
    \label{tab:CrowdPose}
        \vspace{-0.4cm}
\end{table}

\begin{table*}[h]
    \begin{center}
        \begin{minipage}{0.35\linewidth}
        \centering
        \resizebox{\linewidth}{!}{
            \makeatletter\def\@captype{table}\makeatother
        \begin{tabular}{cc|cccc}
            Pose Error &  Loop  & AP  & AP$_{M}$ & AP$_{L}$ & Epoch\\
            \shline
             &  & 70.9   & 65.2 & 79.2 & 60e \\
            \checkmark &  & 72.1  & 66.5 & 80.3 & 45e \\
            \checkmark & \checkmark &  \textbf{73.0}  & \textbf{68.1} & \textbf{80.5} & \textbf{40e} \\
            \shline
        \end{tabular}
            }
                     \vspace{-0.2cm}
                \caption{Impact on \textbf{key components} of \modelname-C0.}
            \label{tab:each_comp}
        \end{minipage}
        \hspace{-0.3cm}
        \quad
                \begin{minipage}{0.29\linewidth}
        \centering
        \resizebox{\linewidth}{!}{
            \makeatletter\def\@captype{table}\makeatother
        \begin{tabular}{l|cccccccc}
            Strategies  & C2  & C4 & C6 & C8  \\
            \shline
            Random  & 84.2  & 90.1 & 94.1 & 96.5 \\
            Low score  & 88.5 & 93.0  & 95.6 & 97.6 \\            
            Worse &   \textbf{90.3} & \textbf{96.4}  & \textbf{98.1} & \textbf{98.7}\\
            \shline
        \end{tabular}
            }
            \vspace{-0.2cm}
    \caption{Ablation study on three \textbf{click strategies}.}
            \label{tab:click_strategy}
        \end{minipage}
        \hspace{-0.3cm}
        \quad
        \begin{minipage}{0.33\linewidth}
        \centering
        \resizebox{\linewidth}{!}{
            \makeatletter\def\@captype{table}\makeatother
        \begin{tabular}{l|ccccc}
            Strategies  & C2 & C4 & C6 & C8  \\
            \shline 
            Only Once & 90.1 & 95.2  & 97.4 & 98.6  \\
            Progressive & \textbf{90.3} & \textbf{96.4} & \textbf{98.1} & \textbf{98.7}\\
            \shline
        \end{tabular}
            }
                
                \caption{Ablation study on two \textbf{loop strategies}.}
            \label{tab:loop_strategy}
        \end{minipage}
    \end{center}
\vspace{-0.3cm}
\end{table*}

\subsection{Annotation Comparisons}
We investigate the advantages of \modelname in different annotation scenes, \textit{i.e.,} in-domain natural scene (COCO), out-of-domain artificial scene (Human-Art), and crowded scenes (OCHuman and CrowdPose). The compared pose estimators comprehensively include top-down (TP), bottom-up (BU), and one-stage (OS) models.

\textbf{Time Cost Comparisons.} In Tab.~\ref{tab:user}, we compared \modelname with two other annotation schemes, one using manual-only annotation, and the other using ViTPose~\cite{xu2022vitpose} to detect initial predictions, followed by manual correction for incorrect keypoints. We conduct a study where ten users annotate the same ten images (which proved challenging for direct prediction via various methods) using different strategies, and we calculate the average and variance time it took to annotate a single person. The results show \modelname significantly reduces the annotation time cost, especially in the out-of-domain annotation scene, achieving a speedup of 10 times compared to manual-only annotation and 5 times compared to the SOTA model with manual correction.

\textbf{NoC Metric Comparisons.} Tab.~\ref{tab:NoC} shows the performance of \modelname and ViTPose in terms of the NoC metric. Our results demonstrate that \modelname with a much smaller backbone can require fewer human corrections to achieve different AP requirements compared to ViTPose, reducing manual effort by $31.4\%$ and $36.3\%$ when the target AP is set to $95$ (NoC$@$95) for COCO and Human-Art, respectively. Non-interactive deep models tend to suffer from model bias and fail on OOD annotation scenes, while \modelname can significantly mitigate this problem.

\begin{table}[h]
    \begin{center}
    \resizebox{\linewidth}{!}
    {%
        \begin{tabular}{l|ccc|c}
            Method  & AP@21 & AP@17 & AP@4 & Correctable Range \\
                        \shline
            \multicolumn{5}{l}{\textit{\cellcolor{Gray}Training on COCO}} \\
                    \modelname-C0 & 25.1 & 40.5  & 0 & -\\
            \shline
                         \multicolumn{5}{l}{\textit{\cellcolor{Gray}Interactive training with \textbf{100} annotated images}} \\
            \modelname-C0 & 47.1{\color{Red}$\uparrow_{22.0}$} & 52.0{\color{Red}$\uparrow_{11.5}$} & 29.1{\color{Red}$\uparrow_{29.1}$} & - \\
            \modelname-C2 & 58.2 ({\color{Red}{+5.4}}) & 64.8 ({\color{Red}{+5.6}}) & 33.2 ({\color{Red}{+4.1}})&  1-17 \\ 
            \modelname-C2 & 59.0 ({\color{Red}{+3.7}}) & 61.1 ({\color{Red}{+3.3}}) & 56.3 ({\color{Red}{+8.9}}) & 1-21\\ 

            \shline
             \multicolumn{5}{l}{\textit{\cellcolor{Gray}Interactive training with \textbf{1000} annotated images}} \\
            \modelname-C0 & 55.0{\color{Red}$\uparrow_{29.9}$} & 58.8{\color{Red}$\uparrow_{18.3}$} & 40.9{\color{Red}$\uparrow_{40.9}$} & - \\
            \modelname-C2 & 69.2 ({\color{Red}{+6.4}}) & \textbf{74.7}  ({\color{Red}{+6.8}}) &45.4 ({\color{Red}{+4.5}}) & 1-17\\ 
            \modelname-C2 & \textbf{70.4} ({\color{Red}{+5.9}}) & 71.0 ({\color{Red}{+5.5}}) & \textbf{67.1} ({\color{Red}{+7.5}}) &  1-21 \\ 
            \shline
        \end{tabular}
    }
    \end{center}
    \vspace{-0.5cm}
    \caption{Effect on \textbf{adaptation $17$ to $21$ keypoints} from COCO ($17$ keypoints) to Human-Art ($21$ keypoints).}
    \label{tab:adapt_Human-Art}
        \vspace{-0.5cm}
\end{table}

\subsection{In-domain Keypoint Detection}
We verify the effectiveness of \modelname in comparison to other state-of-the-art methods in the in-domain scene in Tab.~\ref{tab:coco_main} and~\ref{tab:COCO}, where we train our models on COCO \texttt{train} set and validate them on COCO \texttt{val} set.

\textbf{Comparison with \textit{Model+Manual Correction} methods:} We simulate manual correction on the output results by replacing worse keypoints with ground-truth. The results show that \modelname can achieve better performance compared to ED-Pose, Poseur and ViTPose, with the same amount of human effort. 
For instance, when modifying $4$ incorrect keypoints per person, \modelname achieves $96.4$ AP, which is $8.1$ AP higher than ViTPose.

\textbf{Comparison with \textit{Model-Only} methods:} \modelname-C0, which does not require user corrections, achieves state-of-the-art results using the same ResNet-50 backbone in a fully end-to-end manner. This remarkable performance is attributed to the effective training facilitated by pose error modeling. Notably, \modelname-C0 outperforms ED-Pose by $1.4$ AP with a faster inference time.

\subsection{Out-of-domain Keypoint Detection}
To demonstrate the generalization ability of \modelname, we further evaluate it in the OOD scene, where we train our models on COCO \texttt{train} set and validate them on Human-Art \texttt{val} set (only $17$ keypoints here) in Tab.~\ref{tab:Human-Art_main} and~\ref{tab:Human-Art}.

\textbf{Comparison with \textit{Model+Manual Correction} methods:} In Tab.~\ref{tab:Human-Art_main} and~\ref{tab:Human-Art}, \modelname demonstrates robust performance in such out-of-domain scenario, outperforming ViTPose by $29.5$ AP, Poseur by $35.2$ AP and ED-Pose by $18.3$ AP when clicking $3$ keypoints per-person. Furthermore, \modelname's performance remains consistent across other settings as well.

\textbf{Comparison with \textit{Model-Only} methods:} \modelname-C0 outperforms all two-stage or one-stage approaches, significantly surpassing the SOTA ViTPose model by $11.8$ AP. It also achieves an improvement of $3.0$ AP over ED-Pose.

\subsection{Crowded Scene Keypoint Detection}
Tab.~\ref{tab:OCHuman} and~\ref{tab:CrowdPose} investigates the effectiveness of \modelname in the crowded scene. Here, we compare our \modelname to the baseline model ED-Pose with the same ResNet-50 backbone. Specifically, \modelname-C0 outperforms ED-Pose by $2.5$ AP on OChuman and $0.7$ AP on CrowdPose in the \textit{model-only} setting, showing its robustness in crowded scenes. 
Compared with \textit{ED-Pose+manual correction}, \modelname exhibits significant improvements when receiving user clicks. This is because it not only adjusts the classification scores for all candidate predictions but also enhances localization accuracy. Both of these enhancements greatly contribute to improving annotation efficiency.

\begin{figure*}[h]
\vspace{-0.3cm}
\begin{center}
    \includegraphics[width=1\textwidth]{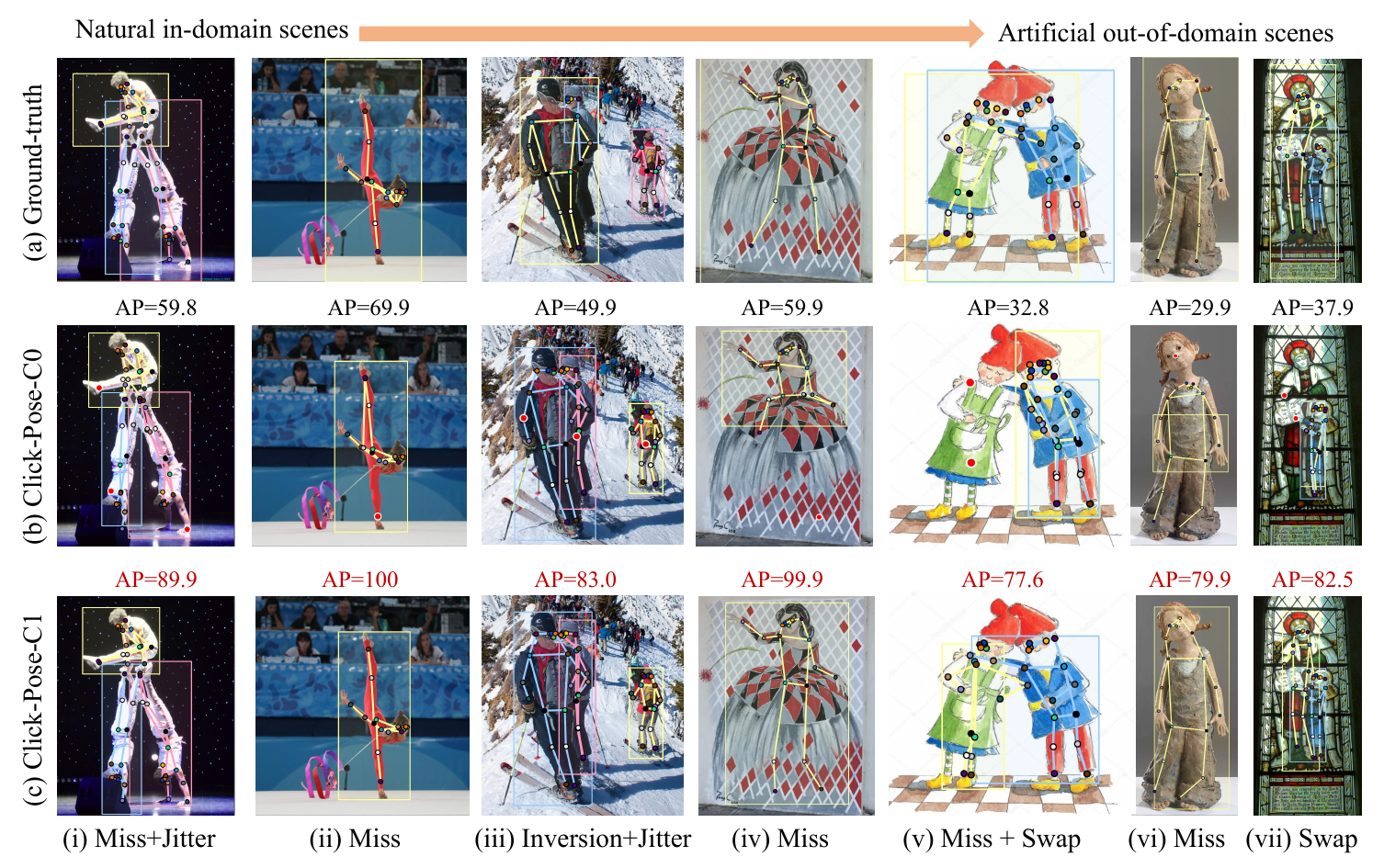}
\end{center}
\vspace{-0.7cm}
\caption{Visualization of the effects of the proposed \modelname with only \textbf{one} user click per person from  in-domain scenes to out-of-domain scenes (directly test on them). The red dots in row (b) represent user clicks for \modelname-C0.}
\vspace{-0.2cm}
\label{figure:click1_viz}
\end{figure*}

\subsection{Ablation Study}

\textbf{Two Key Components.} We evaluate the effectiveness of the proposed two key components on the COCO \texttt{val} set, as shown in Table~\ref{tab:each_comp}. \textbf{First}, \modelname incorporates pose error modeling to enhance the self-correction ability of the model. Our results demonstrate that this training strategy can lead to a $1.2$ AP improvement and reduce the convergence time from $60$ to $40$ epochs compared to the baseline model~\cite{yang2023edpose}.
\textbf{Second}, the human-feedback loop training can also provide an additional improvement of $0.9$ AP by enhancing the model's robustness.

\textbf{Click Strategies.}
In experiments, we take the users to correct a worse keypoint by default. Besides, we explore two other strategies: random clicking and clicking on the keypoint with a low confidence score. 
In Tab.~\ref{tab:click_strategy}, \modelname shows consistent improvement under all three click strategies. Correcting the worse keypoint is the most intuitive annotation way and yields the best performance.

\textbf{Loop Strategies.}
During inference, we set the progressive loop by default, where we only modify one worst keypoint in each loop iteration. We also explore the effectiveness of directly modifying multiple keypoints in a single loop iteration. Tab.~\ref{tab:loop_strategy} gives the performance trends of the two strategies for different numbers of clicks, showing that using the progressive loop strategy can obtain great results, and it is also more intuitive and user-friendly in practice.

\begin{figure*}[h]
\begin{center}
    \includegraphics[width=1\textwidth]{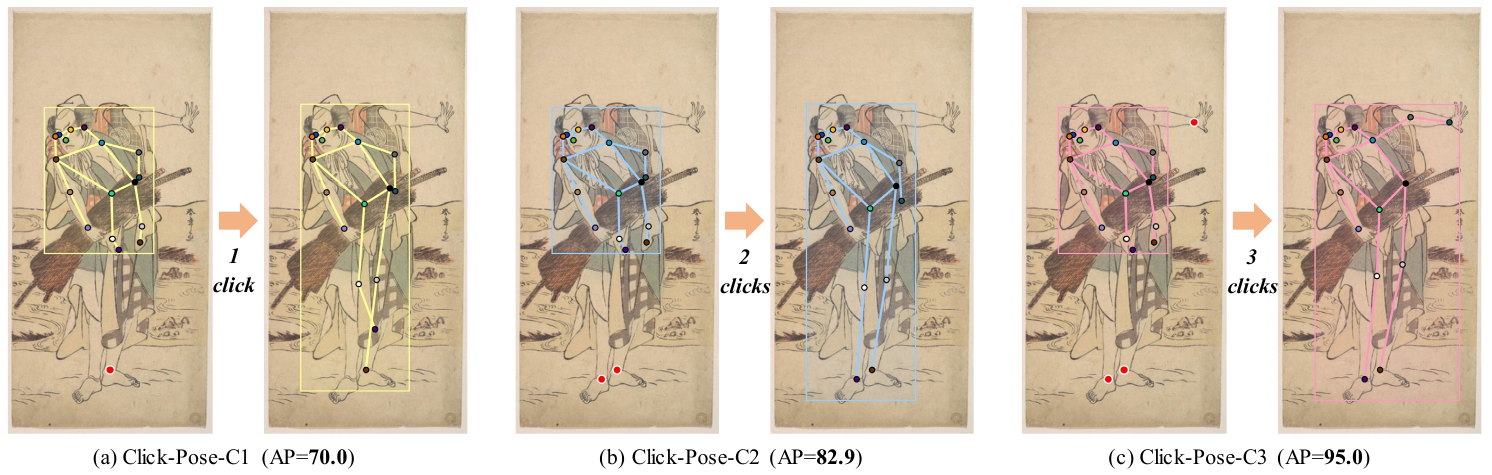}
\end{center}
\vspace{-0.2cm}
\caption{Visualization of the effects of the proposed \modelname with varying numbers of clicks (e.g., one to three). The AP of \modelname-C0 is \textbf{40.0} AP, while \modelname-C3 could refine its results based on three clicks to achieve \textbf{95.0} AP.}
\label{figure:click3_viz}
\vspace{-0.2cm}
\end{figure*}

\subsection{Adaptation to Different Keypoints}
We investigate the adaptability of \modelname in handling additional keypoints that are not included in the original training dataset. Specifically, we train the model using COCO with $17$ labeled keypoints and finetune it on a small set of images (e.g., $100$ and $1000$) in Human-Art labeled with $21$ keypoints, including $4$ additional keypoints that are not defined in COCO.
Tab.~\ref{tab:adapt_Human-Art} reports AP$@21$ for all $21$ keypoints, AP$@17$ for $17$ keypoints defined by COCO, and AP$@4$ for $4$ additional keypoints that require adaptation. Moreover, we provide two options for the range of correctable keypoints, \textit{i.e.,} $1$-$17$ and $1$-$21$. The results demonstrate that when the user corrects keypoints within the $17$ keypoints defined by COCO, \modelname can refine an additional $4$ keypoints. Furthermore, when expanding the range of correctable keypoints to include all $21$ keypoints, the AP$@4$ score is further improved. Importantly, \modelname with limited annotated images can improve 22.0 AP to about 30.0 AP for AP$@21$ without manual correction. 

\subsection{Qualitative Results}
Fig.~\ref{figure:click1_viz} illustrates the effectiveness of the proposed \modelname in both natural in-domain and artificial out-of-domain scenes when receiving only \emph{one} user click.  By leveraging the human feedback loop, \modelname can refine other incorrect keypoints and boxes with user interaction. We show four typical pose error corrections, indicating the effectiveness and efficiency of our proposed method. 
Moreover, Fig.~\ref{figure:click3_viz} shows how \modelname achieves increasingly better results with increasing clicks in challenging scenarios. As the number of clicks increases from $1$ to $3$, the AP score dramatically increases from $40$ to $95$.
\section{Conclusion and Future Work}
\textbf{Conclusion.} This work proposes a novel interactive keypoint detection task incorporating a human-in-the-loop strategy and presents \modelname, an end-to-end neural interactive keypoint detector. \modelname introduces two key components: a pose error modeling scheme and an interactive human-feedback loop. By effectively combining the model with user clicks, \modelname reduces labeling costs by over ten times compared to manual annotation. We hope this work will benefit the community by highlighting the importance of interaction between models and users.

\textbf{Future Work.} This work mainly focuses on multi-person 2D human pose estimation. There are some potential directions for future work.
(I) \textbf{Interactive Whole-body Annotation}: Our work simply considers the mainstream $17$ or $21$ body keypoints. 
When dealing with more complex and dense keypoints (e.g., $133$ keypoints~\cite{yang2023effective,Jin2020WholeBodyHP}), annotating small and blurry areas, like hands and faces, presents greater challenges.
Importantly, these densely labeled body parts often exhibit locally structured spatial relationships that can be leveraged, making the task of labeling dense keypoints quite promising.
(II) \textbf{Interactive Multi-task Annotation}: Our work has focused on annotating human body keypoints and their potential assistance in annotating body boxes (please see supplementary material). Similar to recent SAM~\cite{kirillov2023segany}, a more exciting direction is combining annotations from different tasks (like 2D/3D pose estimation, body parsing, and textual descriptions). A unified model could extract shared features and use different branches to obtain user inputs and estimate various annotations. Changing one annotation could affect others, offering a versatile and comprehensive annotation approach.
(III) \textbf{Interactive 3D Annotation}: Annotating 3D needs high-cost devices and complex processing. Could we annotate 3D information (e.g., point cloud, mesh, keypoints) in the 2D space effectively~\cite{valentin2015semanticpaint,kontogianni2023interactive,shen2020interactive}? This is an intriguing opportunity to expand this approach into the 3D domains. 

\section*{Acknowledgment}
The work is partially supported by the Young Scientists Fund of the National Natural Science Foundation of China under grant No.62106154, by the Natural Science Foundation of Guangdong Province, China (General Program) under grant No.2022A1515011524, and by Shenzhen Science and Technology Program JCYJ20220818103001002 and by Shenzhen Science and Technology Program ZDSYS20211021111415025.
{\small
\bibliographystyle{ieee_fullname}
\bibliography{main}
}

\end{document}


\title{---Supplementary Material---\\Neural Interactive Keypoint Detection}

\author{Jie Yang$^{1,2}$\thanks{Work done during an internship at IDEA.}~~, Ailing Zeng$^{1}$\thanks{Corresponding author.}~~, Feng Li$^{1}$, Shilong Liu$^{1}$, Ruimao Zhang$^{2}$\footnote[2]{}~~, Lei Zhang$^{1}$ \\
$^{1}$International Digital Economy Academy\\
$^{2}$School of Data Science, Shenzhen Research Institute of Big Data, \\The Chinese University of Hong Kong, Shenzhen \\
\texttt{\small{\{jieyang5@link,zhangruimao@\}cuhk.edu.cn}}\\
\texttt{\small{\{zengailing,lifeng,liushilong,leizhang\}@idea.edu.cn}}\\
\url{https://github.com/IDEA-Research/Click-Pose}
}

\ificcvfinal\thispagestyle{empty}\fi

\maketitle
\section*{Overview}
This supplementary material presents more details and additional results not included in the main paper due to page limitation. The list of items included are:

\begin{itemize}
    \item Inference time comparison between the whole model and the only decoder in Sec.~\ref{sec:infere}.
    \item Additional experimental analyses on human detection benefits in Sec.~\ref{sec:supp_bene}.
    \item Discussion the society impacts in Sec.~\ref{sec:supp_so}.
\end{itemize}

\section{Inference Time Comparison}
\label{sec:infere}
We compare the time it takes for \modelname-C0 to generate prediction results with the time required for decoder loop refinement. The results presented in Tab.~\ref{tab:infere} demonstrate that receiving user feedback at the decoder is more efficient than at the input image.
\begin{table}[h]
    \begin{center}
    \scriptsize
        \begin{tabular}{l|c}
          Methods  & Inference time [ms] \\
            \midrule
            \modelname-C0 & 48\\
            Decoder Loop & \textbf{12}\\
            \bottomrule
        \end{tabular}
    \end{center}
            \vspace{-5mm}
    \caption{Comparisons of \textbf{the inference time}.}
            \vspace{-5mm}
    \label{tab:infere}
\end{table}

\section{Benefit to Human Detection}
\label{sec:supp_bene}
Besides the significant improvements in human keypoint detection tasks, as shown in the main paper, the proposed interactive human-feedback loop in \modelname can also help to adjust the positions of human boxes for better human detection. We use the maximum and minimum values of refined keypoints for each person to regularize the width and length of the box. We take the three clicks (C3) and five clicks (C5) as examples, Tab.~\ref{tab:human_dete} demonstrates that 
\modelname can consistently improve $AP_M$ and $AP_L$ on both in- and out-of-domain datasets.

\begin{table}[h]
    \begin{center}
    \scriptsize
        \begin{tabular}{l|ccc|ccc}
            Click & None & C3 & C5  & None & C3 & C5  \\
            \midrule
                    & \multicolumn{3}{l}{\textit{\cellcolor{Gray}COCO \texttt{val}}} &  \multicolumn{3}{l}{\textit{\cellcolor{Gray}Human-Art \texttt{val}}}   \\
            AP$_{M}$ & 68.6  & 69.8 &  \textbf{70.3} & 3.7  & 8.2 & \textbf{9.2}\\
            AP$_{L}$ & 79.0  & 80.0  &  \textbf{80.4} & 14.6  & 21.3 & \textbf{22.8}\\
            \bottomrule
        \end{tabular}
    \end{center}
        \vspace{-5mm}
    \caption{The \modelname's impact on \textbf{Human Detection}.}%
    \vspace{-5mm}
    \label{tab:human_dete}
\end{table}


\section{Society Impact}
\label{sec:supp_so}
With the development of deep learning, the ability of large models in many fields has almost reached the level of human knowledge, especially in creative generation types of work (e.g., recent Stable diffusion models on image generation), which has developed much faster than expected. 
While humans often worry and fear that machines will replace humans and the work they are doing, this article, by exploring how better collaboration with humans on interactive keypoint annotation can reduce repetitive manual efforts and increase the interactive fun of the work, allowing humans to be more productive and spend more time on making critical decisions.
Meanwhile, this work explores the problems of the model bias and performance bottleneck, which makes humans/annotators/users essential to high-quality annotations. Introducing human feedback in an interactive way can improve the usability, transparency, and trustworthiness of deep models. This will help to enable better human-machine interaction and a wider range of human-centric applications, like healthcare, avatar creation, and autonomous robotics. 
At last, using a fast and efficient annotation method can create more high-quality data, making existing large models greater continuously. The data and model will boost each other significantly.